\PassOptionsToPackage{greek,english}{babel}
\documentclass[a4paper,UKenglish]{lipics-v2016}
\usepackage{microtype}
\usepackage{color}
\usepackage{textgreek}

\usepackage[most]{tcolorbox}

\usepackage{xparse}

\def\exampletext{Example} 

\NewDocumentEnvironment{testexample}{ O{} }
{
\colorlet{colexam}{green!35!black} 
\newtcolorbox[use counter=testexample]{testexamplebox}{%
    empty,
    title={\exampletext: #1},
    attach boxed title to top left,
       minipage boxed title,
    boxed title style={empty,size=minimal,toprule=0pt,top=4pt,left=3mm,overlay={}},
    coltitle=colexam,fonttitle=\bfseries,
    before=\par\medskip\noindent,parbox=false,boxsep=0pt,left=3mm,right=0mm,top=2pt,breakable,pad at break=0mm,
       before upper=\csname @totalleftmargin\endcsname0pt, 
    overlay unbroken={\draw[colexam,line width=.5pt] ([xshift=-0pt]title.north west) -- ([xshift=-0pt]frame.south west); },
    overlay first={\draw[colexam,line width=.5pt] ([xshift=-0pt]title.north west) -- ([xshift=-0pt]frame.south west); },
    overlay middle={\draw[colexam,line width=.5pt] ([xshift=-0pt]frame.north west) -- ([xshift=-0pt]frame.south west); },
    overlay last={\draw[colexam,line width=.5pt] ([xshift=-0pt]frame.north west) -- ([xshift=-0pt]frame.south west); },%
    }
\begin{testexamplebox}}
{\end{testexamplebox}\endlist}
\hypersetup{
pdftitle={Learning Noun Cases Using Sequential Neural Networks},
pdfsubject={Research Proposal},
pdfauthor={Sina Ahmadi},
pdfkeywords={noun declension--neural networks--morphological dependency--multi-task learning}
}
\title{Learning Noun Cases Using Sequential Neural Networks}
\titlerunning{A Proposal to Research}
\author[]{Sina Ahmadi}
\affil[]{M.Sc. in Machine Learning and Natural Language Processing\\Paris Descartes University \texttt{ahmadi.sina@outlook.com}}
\authorrunning{Sina Ahmadi} 
\EventEditors{Sina Ahmadi}
\EventNoEds{2}
\EventLongTitle{}
\EventShortTitle{}
\EventAcronym{}
\EventYear{}
\EventDate{}
\EventLocation{}
\EventLogo{}
\SeriesVolume{}
\begin{document}

\maketitle

\begin{abstract}
Morphological declension, which aims to inflect nouns to indicate number, case and gender, is an important task in natural language processing (NLP). This research proposal seeks to address the degree to which Recurrent Neural Networks (RNNs) are efficient in learning to decline noun cases. Given the challenge of data sparsity in processing morphologically rich languages and also, the flexibility of sentence structures in such languages, we believe that modeling morphological dependencies can improve the performance of neural network models. It is suggested to carry out various experiments to understand the interpretable features that may lead to a better generalization of the learned models on cross-lingual tasks.



\end{abstract}

\section{Introduction}

Over the past decade, neural networks have become the state-of-the-art algorithms in most of the tasks in NLP. The improvements in performance thanks to these algorithms has come at the cost of our understanding of each network. Unlike traditional feature-based classifiers that assign and optimize weights to varieties of human-interpretable features, analyzing the behavior of neural network models is more challenging due to their opaque functionality nature. To remedy this, an empirical research is proposed in this report to encourage neural networks to develop more sophisticated generalizations by learning noun cases. Therefore, we will be able to investigate the degree to which neural network models are sensitive to morphology in language. It also enables us to understand the weaknesses and strengths of each model and to create a set of constraints for enhancing the performance of neural machine translation (NMT) models in cross-lingual tasks.

\section{Background}

The work discussed in this study is closely related to those of  Linzen et al. \cite{linzen2016assessing} and Williams et al. \cite{williams2017learning}.

Linzen et al. investigate the extent to which RNNs learn to model syntactic structures while they do not explicitly incorporate with. To address this problem, they focus on tasks commonly regarded as evidence for syntactic structure in human languages such as subject-verb agreement and number prediction. Then, the capacity of the LSTM architecture of RNN to learn structural dependencies is determined under different supervision conditions. In the number prediction task, for instance, the strongest supervision is considered, while in the language modeling objective no grammatical supervision has been used. The results suggest that this explicit supervision is required for learning the agreement dependency. Besides, it is observed that the language modeling signal is insufficient for capturing syntax-sensitive dependencies, and more direct supervision should be considered if such dependencies need to be obtained.  

In addition to the RNN-based models which have shown remarkable success in different tasks in NLP, TreeRNNs have also been applied to various tasks, particularly to the tasks where syntactic and semantic analysis is essential such as natural language inference or recognizing textual entailment. Relying on this network, Williams et al. re-implement two models (RL-SPINN and ST-Gumbel) based on the latent tree learning model to perform on textual entailment to learn to parse a sentence and then to interpret it. Experiments on the SNLI and MultiNLI corpora suggest that the learned trees perform as well or even better than conventional parser such as Penn Tree Bank. On the other hand, their models tend to produce shallower parses which do not resemble semantic or syntactic grammar formalism.

Although the RNN-based and the tree-based models differ primarily in the ways that they train their components, one based on the sequences and the other based on the hierarchical structure, both of these studies try to investigate the degree to which these models can improve the performance of their specific task. In a broader level, Williams et al. also evaluate the degree to which their neural network learns similar grammar across random initializations. They initially try to investigate the impact of semantics and syntax as two determining features in the performance of the network and also analyze if the learned grammars follow a specific pattern. On the other hand, Linzen et al. emphasize on the LSTM architecture of RNN and its capacity in learning dependencies that are less sensitive to be considered as sequences.

The subject of this research proposal has been addressed in various computational fields from neuroscience to psycholinguistics. Some of the most recent works related to our research plan in interpreting and analyzing neural networks for NLP are summarized as follows: Palangi et al. \cite{DBLPPalangiSHD17} propose modifications over neural network architectures to make them more interpretable, Shen et al. \cite{shen2017neural} present Parsing-Reading-Predict Networks (PRPN) that can simultaneously induce the hierarchical structure and leverage the inferred structure to learn a better language model, Adi et al. \cite{adi2016fine} introduce prediction tasks to analyze information encoded in sentence embeddings about sentence content and word order, Li et al. \cite{DBLPLiCHJ15} focus on text but also visual input for understanding neural networks by visualizing them, and finally Cotterell and Heigold \cite{DBLP170809157} explore a transfer learning scheme for morphological tagging to transfer knowledge from rich languages to the low-resourced ones.

\section{Statement of Problem}
One of the main challenges in processing morphologically rich languages is \emph{data sparsity} due to the abundance of inflected forms for one lexical item. This problem may reduce the performance of neural network models which are typically trained based on probabilistic distributions at various levels. In addition, in most of the morphologically rich languages, cases allow flexibility in word order which adds further complexity in learning them. The following example shows the declension of only one category of neutral nouns in different cases in the Modern Greek. The noun articles are bold and the declined morphemes are specified in blue.

\begin{testexample}[Noun cases in Modern Greek]
\begin{itemize}
\item \textgreek{Πού βρίσκεται αυ\textbf{τό} \textbf{το} κέντρ\textcolor{blue}{\textbf{ο}}.} \{Nominative, singular, neutral\}
\item \textgreek{Ο αέρας \textbf{του} βουν\textcolor{blue}{\textbf{ού}} είναι καθαρός.} \{Genitive, singular, neutral\}
\item \textgreek{Προτιμώ \textbf{το} ποδόσφαιρ\textcolor{blue}{\textbf{ο}} από το κολύμπι.} \{Accusative, singular, neutral\}
\item \textgreek{Πού είναι \textbf{τα} βιβλί\textcolor{blue}{\textbf{α}} σου.} \{Nominative, plural, neutral\}
\item \textgreek{Οι τάξεις \textbf{των} λυκεί\textcolor{blue}{\textbf{ων}} είναι τρεις.} \{Genitive, plural, neutral\}
\item \textgreek{Γνωρίζω δύο προγράμ\textcolor{blue}{\textbf{ματα}} για να χάσω λίγα κιλά.} \{Accusative, plural, neutral\}
\end{itemize} 
\end{testexample}

In this context, we believe that morphological dependency in a phrase may be used to improve the performance of neural network models in learning noun declension. 


\section{Objectives}
In this work, I am interested in methods that help us uncovering a broader variety of patterns of behavior of recurrent networks. Methodologically similar to the work of Linzen et al. \cite{linzen2016assessing}, I would like to investigate the extent to which sequence models can decline noun cases. Given the data sparsity challenge, I would also like to know how to modify the current network architectures or design new networks to reduce the reliance of the models on human annotation and overcome the presented challenges? And finally, by sharing representations and the structures that the network learns, how to provide the models with a better generalization on other tasks, such as machine translation and parsing? Is it possible to define a minimal set of constraints for enhancing the performance of cross-lingual tasks based on the morphological information?

\section{Plan of Action}

\begin{itemize}
\item \textbf{Dataset}: By extracting complete sentences with a specific length containing noun, we create training set (9\%), validation set (1\%) and testing set (90\%). These proportions are determined in a way that the least common constructions be represented as well. Among the various noun cases, we will only focus on accusative, nominative, genitive and dative cases. As a suggestion, we can use the Modern Greek Dependency Treebank \footnote{\url{http://gdt.ilsp.gr/}} as dataset.

\item \textbf{Baseline system}: Given the fact that noun articles occur, in general, before the declined word, we can limit our training instances to the noun and its article and use an $n$-gram model for the baseline system. Although it cannot take the whole words of a phrase into account, it can still provide good results for an initial experiment. 

\item \textbf{Model}: Our model is an RNN-based model with the LSTM gated mechanism. Each word will be encoded as one-hot vectors at character level and then will be embedded in a vector space. By sequentially reading the vector spaces, the prediction of the model is calculated by a softmax classifier in the state layer.
\item \textbf{Evaluation}: The output of the model will be evaluated based on the average BLEU score between the predicted sentences and the
reference ones. We can also use the average Levenshtein distance for calculating word-level accuracy. 
\end{itemize}


Although I delimited the boundary conditions for this research to noun cases, they can be extended to other morphological constructions.

\section{Discussion}
The subject of this research proposal can be addressed by a more extensive range of studies. In addition to the axes as mentioned earlier, I am also interested in probing the ability of the neural networks in learning more structure-sensitive dependencies in human language such as gender agreement, ergativity, acceptability judgment, transformational tasks in pragmatics, negative polarity items, reflexive pronouns, and transduction problem. Evaluating different language models for different languages may also be a compelling task to provide novel techniques.


\bibliography{lipics-v2016-sample-article}

\begin{thebibliography}{1}

\bibitem{linzen2016assessing}
Tal Linzen, Emmanuel Dupoux, and Yoav Goldberg.
\newblock Assessing the ability of lstms to learn syntax-sensitive
  dependencies.
\newblock {\em arXiv preprint arXiv:1611.01368}, 2016.

\bibitem{williams2017learning}
Adina Williams, Andrew Drozdov, and Samuel~R Bowman.
\newblock Learning to parse from a semantic objective: It works. is it syntax?
\newblock {\em arXiv preprint arXiv:1709.01121}, 2017.

\bibitem{DBLPPalangiSHD17}
Hamid Palangi, Paul Smolensky, Xiaodong He, and Li~Deng.
\newblock Deep learning of grammatically-interpretable representations through
  question-answering.
\newblock {\em CoRR}, abs/1705.08432, 2017.

\bibitem{shen2017neural}
Yikang Shen, Zhouhan Lin, Chin-Wei Huang, and Aaron Courville.
\newblock Neural language modeling by jointly learning syntax and lexicon.
\newblock {\em arXiv preprint arXiv:1711.02013}, 2017.

\bibitem{adi2016fine}
Yossi Adi, Einat Kermany, Yonatan Belinkov, Ofer Lavi, and Yoav Goldberg.
\newblock Fine-grained analysis of sentence embeddings using auxiliary
  prediction tasks.
\newblock {\em arXiv preprint arXiv:1608.04207}, 2016.

\bibitem{DBLPLiCHJ15}
Jiwei Li, Xinlei Chen, Eduard~H. Hovy, and Dan Jurafsky.
\newblock Visualizing and understanding neural models in {NLP}.
\newblock {\em CoRR}, abs/1506.01066, 2015.

\bibitem{DBLP170809157}
Ryan Cotterell and Georg Heigold.
\newblock Cross-lingual, character-level neural morphological tagging.
\newblock {\em CoRR}, abs/1708.09157, 2017.

\end{thebibliography}

\end{document}